\documentclass[conference]{IEEEtran}
\IEEEoverridecommandlockouts
\usepackage{cite}
\usepackage{amsmath,amssymb,amsfonts}
\usepackage{algorithm}
\usepackage{algorithmic}
\usepackage{graphicx}
\usepackage{textcomp}
\usepackage{xcolor}
\def\BibTeX{{\rm B\kern-.05em{\sc i\kern-.025em b}\kern-.08em
    T\kern-.1667em\lower.7ex\hbox{E}\kern-.125emX}}
\usepackage{setspace}
\usepackage{enumitem}
\usepackage{multirow}
\usepackage{tikz}
\usepackage{comment}
\usepackage{cleveref}
\usepackage{booktabs}
\usepackage{adjustbox}
\newcommand*\circled[1]{\tikz[baseline=(char.base)]{
		\node[shape=circle,draw,inner sep=0.2pt] (char) {#1};}}
\newcommand*\circledB[1]{\tikz[baseline=(char.base)]{
        \node[shape=circle,fill,inner sep=0.2pt] (char) {\textcolor{white}{#1}};}}
\usetikzlibrary{shapes.geometric}

\usepackage{geometry} 
\usepackage{fancyhdr}
\fancypagestyle{firstpage}{
  \fancyhf{}
  \chead{Accepted at the Design, Automation and Test in Europe Conference (DATE), April 20th - 22nd, 2026 in Verona, Italy.}
  \cfoot{\thepage}
}
\begin{document}



\title{Focus Session: Hardware and Software Techniques for Accelerating Multimodal Foundation Models
\vspace{-0.4cm}
}

\author{\IEEEauthorblockN{Muhammad Shafique, Abdul Basit, Muhammad Abdullah Hanif, Alberto Marchisio, \\ Rachmad Vidya Wicaksana Putra, Minghao Shao}
\IEEEauthorblockA{\textit{eBRAIN Lab, New York University (NYU) Abu Dhabi, Abu Dhabi, United Arab Emirates} \\
\{muhammad.shafique, ab7441, mh6117, alberto.marchisio,  rachmad.putra, shao.minghao\}@nyu.edu}
\vspace{-1cm}
}

\maketitle
\pagestyle{plain}
\thispagestyle{firstpage}


\begin{spacing}{0.94}

\begin{abstract}
This work presents a multi-layered methodology for efficiently accelerating multimodal foundation models (MFMs). 
It combines hardware and software co-design of transformer blocks with an optimization pipeline that reduces computational and memory requirements. 
During model development, it employs performance enhancements through fine-tuning for domain-specific adaptation. 
Our methodology further incorporates hardware and software techniques for optimizing MFMs. 
Specifically, it employs MFM compression using hierarchy-aware mixed-precision quantization and structural pruning for transformer blocks and MLP channels. 
It also optimizes operations through speculative decoding, model cascading that routes queries through a small-to-large cascade and uses lightweight self-tests to determine when to escalate to larger models, as well as co-optimization of sequence length, visual resolution \& stride, and graph-level operator fusion. 
To efficiently execute the model, the processing dataflow is optimized based on the underlying hardware architecture together with memory-efficient attention to meet on-chip bandwidth and latency budgets. 
To support this, a specialized hardware accelerator for the transformer workloads is employed, which can be developed through expert design or an LLM-aided design approach. 
We demonstrate the effectiveness of the proposed methodology on medical-MFMs and on code generation tasks, and conclude with extensions toward energy-efficient spiking-MFMs.
\end{abstract}


\section{Introduction}

The rapid advancements of artificial intelligence (AI) have led to the emergence of \textit{foundation models}, i.e., models that are trained with massive datasets to learn diverse input representations and can be adapted for domain-specific (downstream) tasks with minimal additional training~\cite{Ref_Xu_SurveyEfficientFMs_CSUR25}. 
The growing interest in foundation models is reflected through the active developments of \textit{Large Language Models (LLMs)}, \textit{Vision Transformer Models (ViTs)}, and \textit{Latent Diffusion Models (LDMs)}~\cite{Ref_Xu_SurveyEfficientFMs_CSUR25}\cite{Ref_Xu_SurveyMultimodalFMs_arXiv24}.
Their generality and flexibility have inspired a new approach on how intelligent systems are built and deployed~\cite{Ref_Xu_SurveyMultimodalFMs_arXiv24}.
Recent works explore how these models can be enhanced for processing multimodal data and solving diverse cross-modality tasks (e.g., text-based image retrieval and generation).
These models are known as \textit{Multimodal Foundation Models (MFMs)}~\cite{Ref_Xu_SurveyEfficientFMs_CSUR25}.

As multimodal models fuse multiple modalities (e.g., text, image, and audio), they enable strong performance in diverse application domains (e.g., image and video generation~\cite{Ref_Ramesh_DALLE_ICML21}\cite{Ref_Khachatryan_Text2Video_ICCV23}, segmentation~\cite{Ref_Carion_SAM3_arXiv25},
visual-question answering~\cite{Ref_Liu_LLaVA_NeurIPS23}, healthcare~\cite{elmir2024advancing}, and embodied robotics~\cite{Ref_Kawaharazuka_ReviewVLA4Robotics_Access25}. 
However, their scale incurs huge computational, memory/storage, bandwidth, and power/energy costs that hinder efficient training and deployment~\cite{Ref_Xu_SurveyEfficientFMs_CSUR25}, especially for resource-constrained platforms (e.g., edge devices).
Therefore, \textit{optimization techniques in hardware and software domain are essential for accelerating MFMs}.

\subsection{Key Challenges in Accelerating MFMs}
\label{Sec_Challenges}

\begin{figure}[t]
\centering
\includegraphics[width=0.96\linewidth]{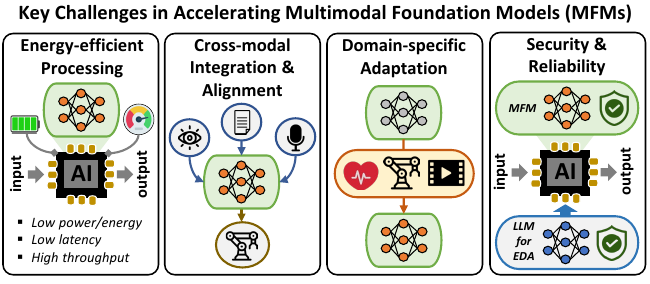}
\vspace{-0.4cm}
\caption{Overview of challenges in accelerating multimodal foundation models.}
\label{Fig_Challenges}
\vspace{-0.5cm}
\end{figure}

We highlight the key challenges in accelerating MFMs in Fig.~\ref{Fig_Challenges} and discuss them in the following.
\begin{itemize}[leftmargin=*]
    \item \textbf{Energy-efficient processing of MFMs:} 
    The size of models leads to substantial computational and memory costs, thus posing energy efficiency challenges in training and inference, especially when they are deployed on resource-constrained platforms. 
    Therefore, model acceleration process should significantly improve its energy efficiency.
    \item \textbf{Cross-modal integration and alignment:} 
    Accelerating the foundation model with heterogeneous modalities (e.g., text, image, and audio) requires an effective multimodal alignment which preserves semantic consistency, while controlling computational and memory overheads.
    \item \textbf{Domain-specific adaptation:} 
    Deployment of the foundation model for a specific domain requires domain-aware fine-tuning and optimization techniques to accelerate the processing and satisfy the constraints of the targeted application (e.g., accuracy, latency, and interpretability). 
    \item \textbf{Security and reliability:} 
    Safety-critical application use cases demand strong robustness to adversarial inputs and distribution shifts, along with safeguards for intellectual property (IP), privacy, and potential data contamination throughout the model lifecycle.
    Therefore, the security and reliability aspects should also be ensured when performing model acceleration. 
\end{itemize}

\begin{figure*}[t]
\centering
\includegraphics[width=0.95\linewidth]{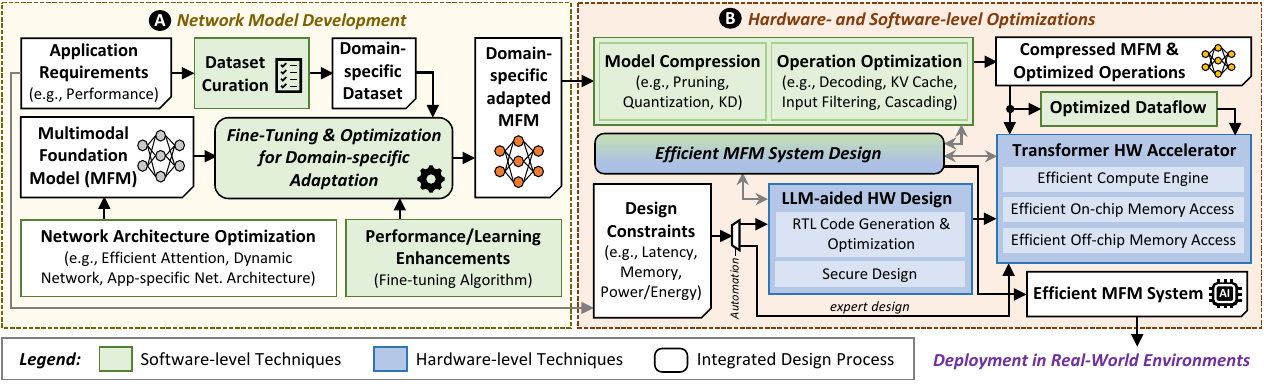}
\vspace{-0.3cm}
\caption{Overview of our multi-layered methodology for accelerating multimodal foundation models (MFMs).}
\label{Fig_OurMethod}
\vspace{-0.5cm}
\end{figure*}

To address each challenge individually, different techniques have been proposed in the literature. 
However, \textit{a methodology for systematically integrating effective hardware and software techniques is still required to enable high-performance and energy-efficient MFM-based systems}.

\subsection{Our Contributions}

In the light of the above discussion, \textbf{the contributions of this paper} are highlighted in the following.
\begin{itemize}[leftmargin=*]
    \item \textit{We present an overview of different key challenges in accelerating MFMs}, including energy efficiency, cross-modal integration \& alignment, domain-specific adaptation, as well as security \& reliability \textbf{(Section~\ref{Sec_Challenges})}.
    \item \textit{We present a multi-layered design methodology for accelerating MFMs}, describing the systematic design pipeline that includes model development for domain-specific adaptation as well as hardware- and software-level optimization techniques (\textbf{Section~\ref{Sec_OurMethod_Overview}}). 
    \item \textit{We present hardware and software techniques for accelerating MFMs}, including model compression, application-driven optimization, transformer hardware accelerator design, and LLM-aided electronic design automation (EDA) that can shorten the hardware design cycle \textbf{(Section~\ref{Sec_OurMethod_SW} and \ref{Sec_OurMethod_HW})}. 
    \item \textit{We extend the energy efficiency of MFMs through spiking neural networks (SNNs)}, developing the suitable optimization techniques for their event-based operations \textbf{(Section~\ref{Sec_OurMethod_Spiking})}. 
\end{itemize}

\section{Our Methodology for Accelerating MFMs}
\label{Sec_OurMethod}

\begin{figure*}[t]
\centering
\includegraphics[width=0.95\linewidth]{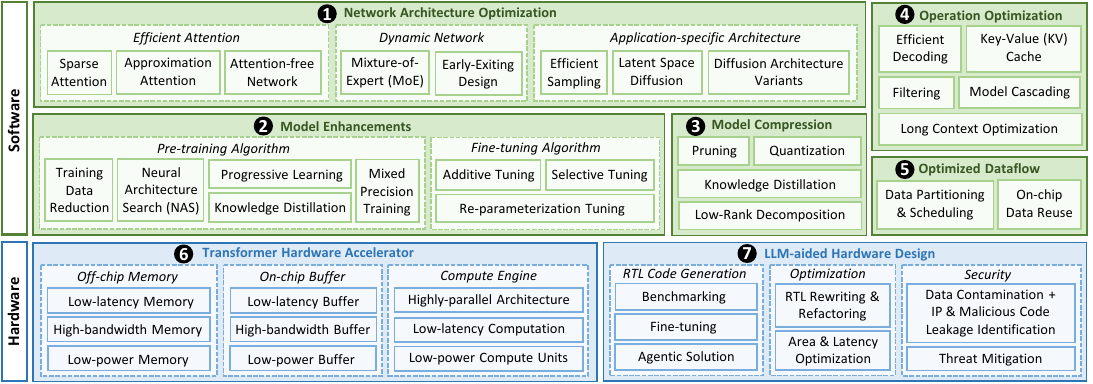}
\vspace{-0.3cm}
\caption{Overview of hardware and software techniques for accelerating MFMs.}
\label{Fig_AllTechniques}
\vspace{-0.5cm}
\end{figure*}

\subsection{Overview}
\label{Sec_OurMethod_Overview}

To enable the acceleration of MFMs while satisfying the application requirements and constraints, we propose \textit{a design methodology that systematically integrates effective hardware- and software-level techniques in a seamless design pipeline}; as shown in Fig.~\ref{Fig_OurMethod} and discussed below.

\smallskip
\circledB{A} \textbf{Network Model Development:}
An MFM typically requires domain-specific adaptation for targeting the downstream tasks. 
Hence, it is important to perform \textit{fine-tuning and optimization for domain-specific adaptation}.
To support this, the following development aspects are essential.
\begin{itemize}[leftmargin=*]
    \item \textit{Dataset curation:} 
    A domain-specific dataset has to be carefully curated and then employed for fine-tuning the model considering the application requirements (e.g., performance). 
    Here, dataset distillation~\cite{Ref_Wang_DatasetDistill_arXiv18} can be employed to reduce the size of training data while preserving the information.  
    \item \textit{Performance/learning enhancements:}
    To improve the performance of the model, pre-training and fine-tuning algorithms can be employed; see \circledB{2} in Fig.~\ref{Fig_AllTechniques}. 
    Pre-training algorithm is used for model development in pre-training phase; while fine-tuning algorithm aims at enhancing performance of the pre-trained model, such as additive tuning~\cite{Ref_Chen_MPrompt_EMNLP23}\cite{Ref_Li_PrefixProp_ACL23}, selective tuning~\cite{Ref_Fu_SAM_AAAI23}\cite{Ref_Huang_GreenAI_ICLR24}, and re-parameterization tuning~\cite{Ref_He_EfficientDM_ICLR24}\cite{hu2021LoRa}.
\end{itemize}
In the case when model development from scratch is needed, we can benefit from optimization techniques for network architecture design, such as efficient attention (e.g., sparse~\cite{Ref_Li_TDANet_ICLR23}, 
approximate~\cite{Ref_Ma_Mega_ICLR23}, 
and attention-free~\cite{Ref_Peng_RWKV_EMNLP23}\cite{Ref_Gu_Mamba_COLM24}), 
dynamic network (e.g., Mixture-of-Expert (MoE)~\cite{Ref_Riquelme_VMOE_NeurIPS21} and 
early-exiting~\cite{Ref_Bae_FREE_EMNLP23}), 
and application-specific architecture~\cite{Ref_Song_DDIM_ICLR21, Ref_Rombach_LDM_CVPR22,Ref_He_ScaleCrafter_ICML23}; see \circledB{1} in Fig.~\ref{Fig_AllTechniques}.

\smallskip
\circledB{B} \textbf{Hardware- and Software-level Optimizations:}
Once the domain-adapted MFM is obtained, further optimizations in hardware and software domains can be performed, as follows. 
\begin{itemize}[leftmargin=*]
    \item \textit{Model compression:}
    The domain-adapted MFM can be compressed through software techniques, such as pruning (i.e., unstructured~\cite{Ref_Frantar_SparseGPT_ICML23, Ref_Sun_Wanda_ICLR24, Ref_Zhang_PlugPlay_ICLR24} and structured~\cite{Ref_Ma_LLMpruner_NeurIPS23, Ref_Song_SLEB_ICML24, Ref_Xia_ShearedLLaMA_ICLR24}), quantization~\cite{Ref_Xiao_SmoothQuant_ICML23, Ref_Lin_AWQ_MLSys24, Ref_Shao_OmniQuant_ICLR24, Ref_Liu_SpinQuant_ICLR25}, knowledge distillation (KD)~\cite{Ref_Gu_MiniLLM_ICLR24}, and 
    low-rank decomposition (LoRD)~\cite{Ref_Li_Losparse_ICML23}\cite{Ref_Kaushal_LoRD_ICMLW24}; see \circledB{3} in Fig.~\ref{Fig_AllTechniques}. 
    Pruning eliminates insignificant weights; 
    quantization lowers the bit-precision of weights and/or activations. 
    KD transfers knowledge from a bigger model (teacher) to a smaller model (student). 
    Meanwhile, LoRD decomposes the weight matrix in the models into multiple smaller matrices.
    Furthermore, there is significant potential in combining multiple compression techniques to achieve higher compression ratios with minimal accuracy loss. 
    For instance, Optimal Brain Restoration (OBR) framework~\cite{Ref_Guo_OBR_arXiv25} employs a training-free approach that jointly leverages pruning and quantization. 
    \item \textit{Operation optimization:}
    Operations of the domain-adapted MFM can be optimized using software techniques, e.g., Key-Value (KV) cache~\cite{Ref_Liu_Scissorhands_NeurIPS23}, efficient decoding (e.g., speculative decoding)~\cite{Ref_Sun_SpecTr_NeurIPS23}, input filtering (i.e., prompt compression~\cite{Ref_Jiang_LLMLingua_EMNLP23} and token pruning~\cite{Ref_Anagnostidis_ContextPruning_NeurIPS23}), model cascading~\cite{Ref_Chen_ModelCascading_IJCNN25}, and long-context optimization~\cite{Ref_Xiao_StreamingLM_ICLR24}; see \circledB{4} in Fig.~\ref{Fig_AllTechniques}.
    KV cache avoids re-computation of KV by reusing the existing results. 
    Speculative decoding avoids costly sequential decoding by using a small model to guess next tokens and a larger model to verify them in a single forward pass.
    Input filtering reduces the input size using prompt compression and token pruning.
    Model cascading routes queries through a small-to-large cascade and uses lightweight self-tests to determine when to escalate to larger models.
    Meanwhile, long-context optimization handles long input via recurrent structure and attention enhancements.   
    \item \textit{Optimized dataflow:} 
    To efficiently execute the compressed MFM and its operations, dataflow should be optimized based on the underlying hardware architecture~\cite{Ref_Putra_ROMANet_TVLSI21}. 
    This execution is scheduled for maximizing data reuse on-chip, since on-chip operations incur significantly lower energy consumption than off-chip memory access~\cite{Ref_Putra_DRMap_DAC20}; see \circledB{5} in Fig.~\ref{Fig_AllTechniques}. 
    \item \textit{Hardware accelerator:} 
    To maximize the efficiency gains in accelerating MFMs, specialized hardware accelerators are employed~\cite{Ref_Li_LLMhwSurvey_arXiv24}; see \circledB{6} in Fig.~\ref{Fig_AllTechniques}
    These accelerators aim to satisfy computational and memory requirements of targeted models. 
    To achieve this, the compute engine can be designed with highly-parallel architecture, low-latency computation, and low-power compute units. 
    Meanwhile, the off-chip and on-chip memories can be designed with low-latency, high-bandwidth, and low-power memory technologies. 
    \item \textit{LLM-aided hardware design:} 
    To enable a fast and scalable hardware accelerator design process, LLMs can be employed to generate register-transfer level (RTL) codes of the accelerator from natural language specifications (including design constraints) in a close-loop design flow (i.e., specifications $\rightarrow$ RTL generation $\rightarrow$ verification $\rightarrow$ synthesis $\rightarrow$ feedback); see \circledB{7} in Fig.~\ref{Fig_AllTechniques}.  
    Furthermore, this approach also allows a seamless integration with the existing EDA tools.
\end{itemize}
Once the MFM and hardware accelerator are developed and optimized, they are then integrated as a system for deployment in the targeted real-world application.

\subsection{Software-level Optimization Methods}
\label{Sec_OurMethod_SW}

\subsubsection{\textbf{Model Compression}}

\begin{figure}[t]
\centering
\includegraphics[width=\linewidth]{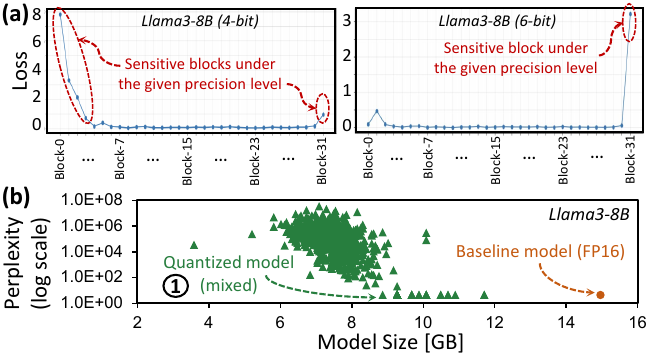}
\vspace{-0.7cm}
\caption{Our experimental results on the WikiText-2 dataset for \textbf{(a)} the impact of different precision levels in each transformer block of Llama3-8B; and \textbf{(b)} the performance of Llama3-8B under weight quantization considering different combinations of precision levels across transformer blocks.}
\label{Fig_Quant4Llama3t8B}
\vspace{-0.3cm}
\end{figure}

MFMs can achieve impressive performance at the cost of huge computational and memory requirements, hence huge power/energy consumption. 
To address this, model compression is essential. 
In the pruning category, 
SparseGPT~\cite{Ref_Frantar_SparseGPT_ICML23} and Wanda~\cite{Ref_Sun_Wanda_ICLR24} target unstructured layer-wise sparsity, 
Plug-and-Play~\cite{Ref_Zhang_PlugPlay_ICLR24} leverages relative importance of weights and activations, 
LLM-Pruner~\cite{Ref_Ma_LLMpruner_NeurIPS23} employs structured group-wise pruning based on gradient information,  
while SLEB~\cite{Ref_Song_SLEB_ICML24} considers transformer block-level pruning.
In the quantization category, SmoothQuant~\cite{Ref_Xiao_SmoothQuant_ICML23} migrates quantization difficulties from activations to weights, AWQ~\cite{Ref_Lin_AWQ_MLSys24} quantizes most of the weights while keeping the salient ones in higher precision, 
OmniQuant~\cite{Ref_Shao_OmniQuant_ICLR24} uses block-wise quantization error minimization, 
while SpinQuant~\cite{Ref_Liu_SpinQuant_ICLR25} uses a rotation matrix to reduce outliers and enhance quantizability.
These techniques usually rely on uniform quantization across transformer blocks, thus leading to suboptimal memory savings. 
Furthermore, our experimental results demonstrate that different blocks may have different sensitivity under quantization, as shown in Fig.~\ref{Fig_Quant4Llama3t8B}(a). 
Based on this observation, we perform exploration to obtain pareto-optimal quantized models that achieve competitive performance with significant memory savings from the baseline through a mixed-precision approach; see Fig.~\ref{Fig_Quant4Llama3t8B}(b). 
For instance, a selected model at label-\circled{1} has perplexity score of 4.73, which is slightly above the baseline model with perplexity score of 4.46, while significantly saving 40.74\% of memory footprint.   

\subsection{Hardware-level Optimization Methods}
\label{Sec_OurMethod_HW}

\subsubsection{\textbf{Transformer Accelerator}}

When executing Transformers on general-purpose processors or GPUs, inference often suffers from excessive latency and power consumption, limiting practical use at the edge. 
Existing ML accelerators primarily target convolutional workloads and dense linear algebra, offering limited native support for Transformer-specific primitives. 
Operations such as multi-head attention, normalization layers, and nonlinear activations are frequently emulated through sequences of basic kernels, leading to inefficiencies and unnecessary data movement. Several prior designs focus on accelerating isolated components, such as matrix multiplications or attention blocks, without addressing the full inference pipeline~\cite{Ham_2020_A3}. 
Moreover, the reliance on floating-point arithmetic significantly increases silicon area, energy usage, and critical path delay. 
To overcome these limitations, an integer-only execution model that uniformly supports all Transformer operations is essential for achieving both high efficiency and low power consumption.

\begin{figure}[h]
\vspace{-0.3cm}
\centering
\includegraphics[width=\linewidth]{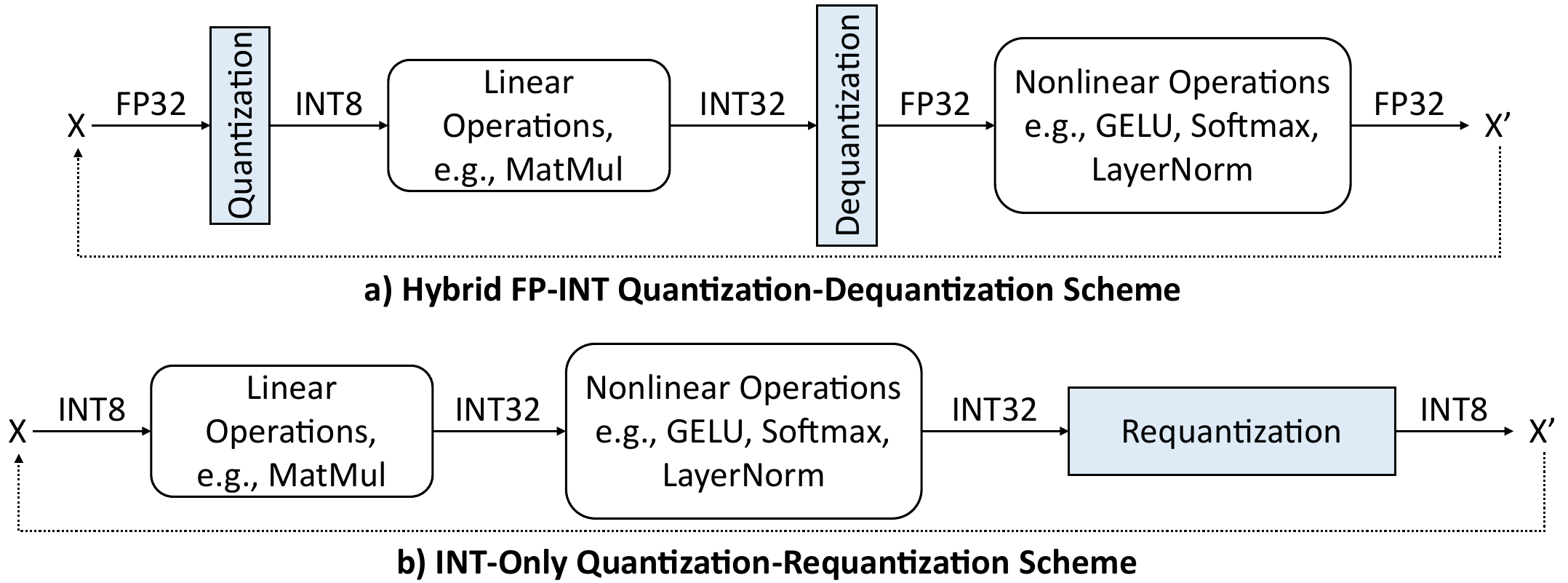}
\vspace{-0.7cm}
\caption{Computation flows for \textbf{(a)}~the Quantization-Dequantization scheme of~\cite{lin2020towards}, and \textbf{(b)}~the quantization-based scheme with \textit{Requantization} blocks~\cite{marchisio2023swifttron}.}
\label{fig:Quantization_dequantization}
\vspace{-0.2cm}
\end{figure}

\smallskip
\textit{Quantization Scheme for Accelerators:}
As shown in Fig.~\ref{fig:Quantization_dequantization}, the accelerator adopts a unified quantization strategy that enables integer-only computation across the entire Transformer pipeline. Model parameters and intermediate activations are represented using 8-bit signed integers, providing a compact and energy-efficient storage format. Intermediate accumulations and nonlinear transformations are carried out in 32-bit integer precision to preserve numerical stability. After each operation, a requantization step rescales the results back to 8-bit precision, ensuring that values remain within a valid dynamic range. This quantization-requantization flow is consistently applied to all major components, including attention, feed-forward layers, normalization, and activation functions, enabling a homogeneous and hardware-friendly design.

\begin{figure}[h]
\vspace{-0.2cm}
\centering
\includegraphics[width=0.98\linewidth]{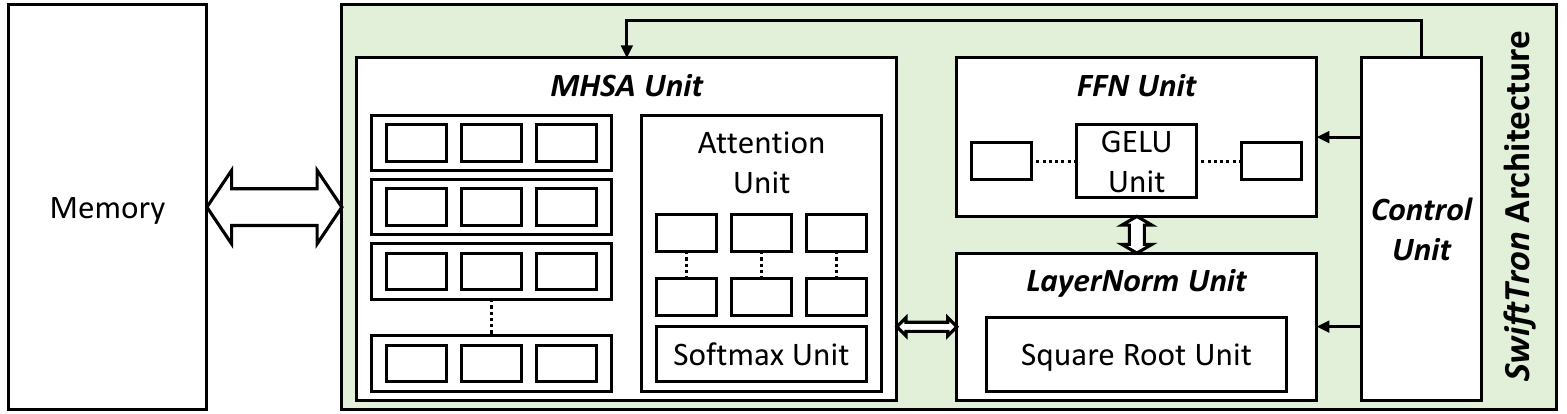}
\vspace{-0.3cm}
\caption{Top-level overview of the \textit{SwiftTron} architecture.}
\label{fig:Top_level_SwiftTron}
\vspace{-0.2cm}
\end{figure}

\smallskip
\textit{Hardware Accelerator Architecture:}
As shown in Fig.~\ref{fig:Top_level_SwiftTron}, the SwiftTron architecture~\cite{marchisio2023swifttron} is organized around a set of tightly coupled functional units optimized for Transformer workloads. 
The matrix multiplication engine serves as the primary compute backbone, handling both projection layers and feed-forward networks with high throughput. 
A dedicated requantization unit follows each compute stage to normalize intermediate results and manage precision transitions efficiently. The multi-head self-attention module orchestrates query, key, and value projections, while the attention engine performs scaled dot-product computations. Specialized units implement Softmax and GELU using integer arithmetic, and a LayerNorm block ensures stable activation distributions. 
On-chip memory buffers minimize external memory access, and a centralized control unit schedules dataflow and synchronizes all components.

An end-to-end integer-only datapath substantially reduces hardware complexity and energy consumption when compared to floating-point alternatives. As shown in \Cref{tab:SwiftTron_speedup}, experimental results demonstrate significant performance gains, achieving over $3.5 \times$ speedups over GPU baselines across language and vision Transformer benchmarks.
The hardware analysis shows that matrix multiplication dominates both area and power consumption (i.e., 55\% and 79\%, respectively), highlighting it as the primary optimization target. 
Overall, the results indicate that a fully integrated, quantized Transformer accelerator can deliver high performance while maintaining a compact and energy-efficient footprint.

\begin{table}[h]
\vspace{-0.3cm}
\centering
\scriptsize
\caption{Accuracy and inference latency comparison between SwiftTron and RTX 2080 Ti GPU.
}
\vspace{-0.2cm}
\label{tab:SwiftTron_speedup}
\begin{adjustbox}{max width=.9\linewidth}
\begin{tabular}{c|c|c|c}
\textbf{Model} & \textbf{Accuracy} & \textbf{Latency} & \textbf{Speedup w.r.t. GPU} \\ \toprule
RoBERTa-base on STT-2 & $95.2\%$ & $1.83\ ms$ & $3.81 \times$ \\ \midrule
RoBERTa-large on STT-2 & $96.4\%$ & $45.70\ ms$ & $3.90 \times$ \\ \midrule
DeiT-S on ImageNet & $79.11\%$ & $1.13\ ms$ & $3.58 \times$
\end{tabular}%
\end{adjustbox}
\vspace{-0.5cm}
\end{table}

\smallskip
\subsubsection{\textbf{LLM-aided Hardware Design}}

We discuss the challenges of current accelerator design, RTL code generation and optimization, as well as security concern in LLM-aided EDA.

\smallskip 
\textit{Challenges of Current Accelerator Design:} 
The design of hardware accelerators for transformer-based models poses major challenges that slow development and deployment. 
Conventional workflows require substantial expertise in digital circuit design and hardware description languages (HDLs), creating high barriers to entry and long design cycles. 
Teams often spend months iterating through specification, RTL implementation, verification, and synthesis, as constraints discovered at later stages force revisions to earlier choices.
Manual design also limits effective exploration of the design space for energy-efficient architectures. 
Engineers rely on prior experience and heuristics, yet the complex interactions among compute, data movement, and on-chip storage make the evaluation of many alternatives impractical. 
These issues are compounded by the rapid evolution of foundation models, where workloads vary widely from dense attention to sparse mixture-of-experts~\cite{kachris2025survey}. 
Therefore, automated hardware generation is increasingly important for translating high-level specifications and constraints into optimized RTL designs with reduced dependence on scarce hardware expertise~\cite{pan2025survey}.

\smallskip
\textit{RTL Code Generation and Optimization:}
LLMs have demonstrated strong ability in generating codes~\cite{shao2024survey} from natural language specifications, providing potential to transform hardware design workflows. This capability spans in multiple levels, including datasets and benchmarks, model-level methods, and agentic architectures, which together enable a closed-loop flow integrating specification, generation, verification, synthesis, and feedback.
High-quality datasets and benchmarks form the foundation of LLM-aided RTL generation. Efforts include VerilogDB~\cite{calzada2025verilogdb} for large-scale training corpora with synthesis validation, while benchmarks such as VerilogEval~\cite{liu2023verilogeval} and RTLLM~\cite{lu2024rtllm} assess functional correctness.
At the model level, domain-adapted LLMs achieve significant gains over general-purpose models through techniques such as domain-adaptive pretraining (e.g., ChipNeMo~\cite{liu2023chipnemo}) and efficient fine-tuning (e.g., RTLCoder~\cite{liu2024rtlcoder}). Closed-loop refinement frameworks such as VeriAssist~\cite{huang2024towards} leverage compiler and simulator feedback for iterative code repair.
At the agentic level, multi-agent systems such as MAGE~\cite{zhao2025mage} coordinate specialized agents for RTL generation, testbench creation, and debugging, reflecting human design workflows. For EDA tool integration, frameworks such as ChatEDA~\cite{wu2024chateda} demonstrate strong task planning, while VeriDispatcher~\cite{wang2025veridispatcher} explores multi-LLM dispatching for RTL generation. For both security and optimization, NetDeTox~\cite{wang2025netdetox} leverages LLMs to minimize chip area overhead while maintaining robustness against GNN-based attacks.

\begin{figure}[h]
\vspace{-0.3cm}
\centering
\includegraphics[width=\linewidth]{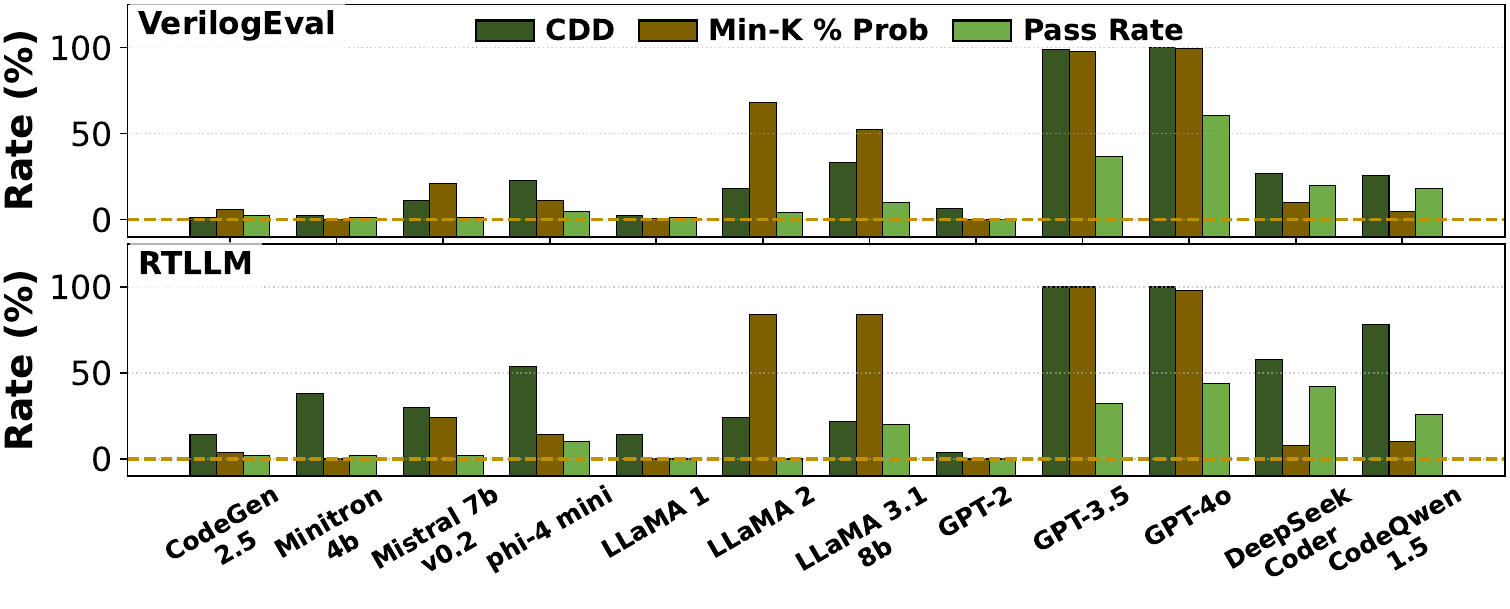}
\vspace{-0.9cm}
\caption{Data contamination rate from pre-training data and pass rate for different LLMs on VerilogEval and RTLLM.}
\label{fig:vericontaminated}
\vspace{-0.3cm}
\end{figure}

\begin{figure}[h]
\vspace{-0.3cm}
\centering
\includegraphics[width=\linewidth]{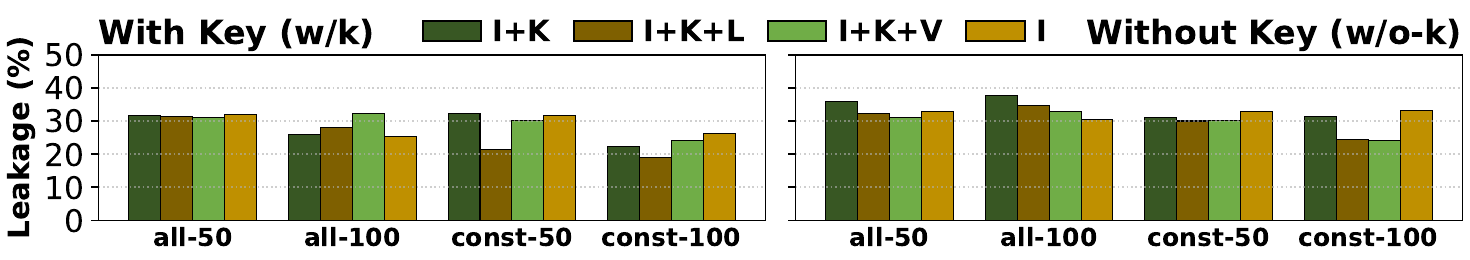}
\vspace{-0.8cm}
\caption{Logic locking (I + K + L and I + K + V) successfully mitigates the potential IP leakage from LLMs.}
\label{fig:verileaky}
\vspace{-0.3cm}
\end{figure}

\smallskip
\textit{Security Concerns in LLM-aided EDA:} 
As LLMs become an integrated part of hardware design workflows, security has emerged as a dual-use research area. 
One direction applies LLMs to hardware security tasks, while the other examines the risks introduced by LLM-assisted design and verification~\cite{wang2024llms}. 
On the offensive side, GHOST shows that an LLM can automate hardware Trojan construction and insertion at the RTL level.
TrojanLoC~\cite{xiao2025trojanloc} further indicates that general-purpose LLMs can help attackers navigate large SoC codebases and identify regions for Trojan insertion or secret leakage. 
On the defensive side, BugWhisperer improves RTL-level vulnerability detection~\cite{tarek2025bugwhisperer}.
LLM-aided EDA also introduces assurance challenges. 
VeriContaminated highlights that benchmark contamination can inflate performance claims and complicate evaluation, and proposes mitigation through token generation probability statistics such as Min-K\% probability combined with similarity checks~\cite{wang2025vericontaminated}. 
Fig.~\ref{fig:vericontaminated} shows different models' contamination rates with two detection approaches against VerilogEval and RTLLM. 
Verileaky shows that models fine-tuned on proprietary RTL may leak IP, and proposes a logic-locking-based approach for mitigation~\cite{wang2025verileaky}, as shown in Fig. ~\ref{fig:verileaky}. 
SALAD further addresses related data risks through machine unlearning to forget contaminated or sensitive RTL patterns~\cite{wang2025salad}.

\section{Case Study in Healthcare and Robotics}

To demonstrate the effectiveness of compression techniques for multimodal large language model (MLLM)-based real-world systems, we present two representative case-studies for different safety-critical domains: (i) medical MLLMs for radiology assistance in low-resource healthcare settings, and (ii) vision–language–action (VLA) pipelines for embodied robotic navigation, which typically demand high energy efficiency due to limited compute, memory, and power budgets.
In both cases, domain-specific adaptation of an MFM is followed by model compression (e.g., quantization and pruning), and hardware-aware deployment on embedded GPUs.

\smallskip
\subsubsection{\textbf{Healthcare VLM}}
We evaluate an optimization and deployment framework for medical MLLMs in which a general-purpose TinyLLaVA-1.5B backbone~\cite{zhou2024tinyllava} is adapted to the biomedical domain and then aggressively quantized.
The optimization stage comprises three fine-tuning phases: (1) biomedical alignment on 600k PMC-15M image--text pairs~\cite{zhang2023biomedclip}, updating only the projection layer; (2) instruction tuning on 60k PMC-15M instruction-style samples to obtain a conversational assistant; and (3) downstream fine-tuning on VQA-RAD, SLAKE, and PathVQA for radiology and pathology VQA.
The resulting TinyLLaVA-Med-F model attains competitive accuracy to larger 7B medical MLLMs with only 1.5B parameters.

\begin{figure}[ht]
\centering
\includegraphics[width=\linewidth]{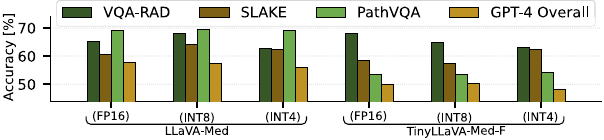}
\vspace{-0.8cm}
\caption{Compression results on medical VQA (closed-question accuracy).}
\label{healthcare}
\vspace{-0.2cm}
\end{figure}

In the compression stage, post-training quantization (PTQ) is applied to both TinyLLaVA-Med-F~\cite{elmir2024advancing} and the 7B LLaVA-Med model~\cite{li2024llavamed}, producing 8-bit and 4-bit variants. Quantization is performed on weights and activations using a calibration set from the medical VQA datasets, without additional gradient updates. This yields a family of low-bit medical MLLMs across model sizes and bit-widths, allowing us to jointly analyze task accuracy, conversational quality, and memory footprint; the key results are summarized in Fig.~\ref{healthcare}.

\vspace{-0.3cm}

VQA accuracy is largely robust to low-bit quantization since both 7B and 1.5B models exhibit only minor variation across datasets, with 8-bit variants occasionally matching or slightly exceeding full-precision performance, consistent with calibration-induced regularization. 
GPT-4 evaluation shows a similar trend, indicating that conversational quality remains stable under quantization. From a deployment perspective, moving to 4-bit precision reduces dynamic memory by $\sim$3$\times$ for the 7B model and $\sim$9$\times$ for the 1.5B model, hence enabling execution on 6GB consumer GPUs and Jetson-class embedded platforms. Together, these results demonstrate that the quantization-centric pipeline achieves substantial memory savings while preserving clinically meaningful performance.

\smallskip
\subsubsection{\textbf{Robotics VLM}}

In robotic perception, we consider open-vocabulary object detection (OVD) as a key \emph{vision–language} capability that underpins embodied navigation.

\vspace{-0.3cm}
\begin{figure}[ht]
    \centering
    \includegraphics[width=\columnwidth]{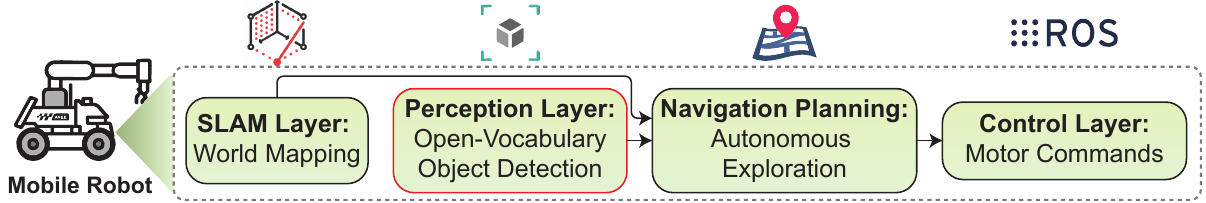}
    \vspace{-0.7cm}
    \caption{Overview of robotics system (SLAM, OVD, embodied navigation).}
    \label{fig:Robotics_CS}
    \vspace{-0.2cm}
\end{figure}

As illustrated in Fig.~\ref{fig:Robotics_CS}, OVD operates at the perception layer, providing semantically grounded object representations that complement geometric state estimation from SLAM and are primarily consumed by navigation and task-planning modules.
OVD outputs (e.g., bounding boxes and language-conditioned labels) inform semantic reasoning, obstacle awareness, and goal specification during planning. Although OVD models do not directly output actions or motor commands, their latency and throughput critically influence the end-to-end responsiveness of embodied robotic systems by enabling faster perception-driven replanning and decision-making~\cite{ROVER}.

OVD models used in robotics span across diverse architectural families, ranging from convolutional detectors augmented with language embeddings for real-time inference (e.g., YOLO-World) to transformer-based vision–language models that rely on global self-attention for stronger semantic generalization (e.g., OWLv2 and GroundingDINO). YOLO-World offer low latency due to convolutional inductive biases and dense prediction heads, but typically trade-off fine-grained semantic alignment and long-tail generalization.
In contrast, transformer-based OVD models achieve superior open-vocabulary accuracy and robustness across diverse categories at substantially higher computational and memory cost. This trade-off motivates the need for systematic compression strategies that can bring these VLMs closer to real-time operation on robotic platforms.

\begin{figure}[ht]
    \centering
    \includegraphics[width=\columnwidth]{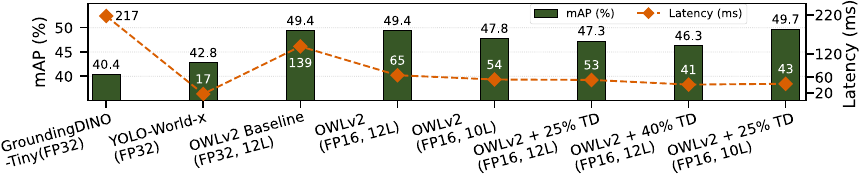}
    \vspace{-0.7cm}
    \caption{OVD performance on COCO val2017 for robotics-relevant perception (TD denotes token drop).}
    \label{fig:OVD}
    \vspace{-0.2cm}
\end{figure}

Starting from this broader OVD landscape, we focus on OWLv2 as a representative transformer-based VLM for robotics perception and investigate a training-free compression pipeline following the multi-layered methodology in Section~II. 
The pipeline combines software-level mixed-precision inference with network-level simplification through encoder layer removal and token sparsification. 
Post-training INT8 quantization is also evaluated but excluded due to its adverse impact on vision–language alignment.
Fig.~\ref{fig:OVD} highlights several key insights.
Mixed-precision inference alone substantially improves throughput while preserving detection accuracy, confirming that OWLv2 representations are robust to reduced numerical precision. Modest structural simplification (i.e., either via layer removal or token drop) yields additional latency reductions with limited accuracy impact, while more aggressive token sparsification exposes a clear accuracy–efficiency trade-off. 
Notably, combining layer removal with moderate token drop recovers most of the baseline accuracy while achieving improved latency profile, significantly narrowing the efficiency gap.

\vspace{-0.2cm}

Beyond COCO, the same compressed configuration generalizes well to LVIS and 13 robotics-focused ODinW datasets. While long-tail categories in LVIS are more sensitive to depth reduction, overall performance remains high.
Across diverse aerial, driving, manipulation, and environmental monitoring scenarios in ODinW, compressed OWLv2 variants consistently achieve multi-fold speedups with comparable or improved F1 scores, indicating that structural pruning and FP16 inference act as effective regularizers and preserve domain generalization for robotics-oriented vision–language perception.

\section{Extensions Toward Spiking-MFMs}
\label{Sec_OurMethod_Spiking}

SNNs have emerged as promising low-power/energy AI algorithms due to their sparse event-based operations~\cite{Ref_Putra_SpikeNAS_TAI25}. 
Therefore, researchers recently leveraged SNNs for MFMs (so-called \textit{Spiking-MFMs}) by applying event-based operations in transformer blocks. 
Here, we discuss the state-of-the-art works, their limitations, and representative optimization techniques. 

\smallskip
\subsubsection{\textbf{Spiking Language Models (SLMs)}}
\begin{figure}[h]
\vspace{-0.2cm}
\centering
\includegraphics[width=\linewidth]{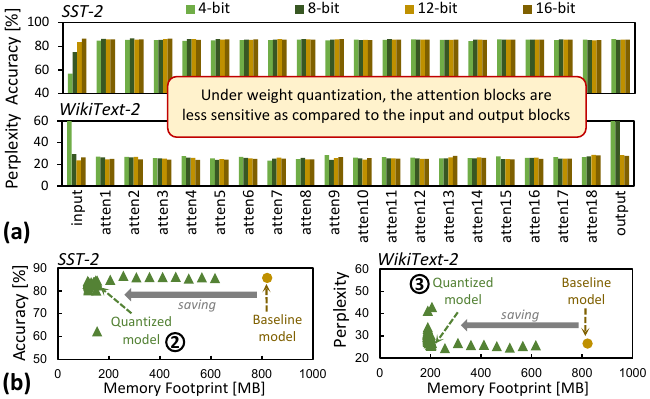}
\vspace{-0.7cm}
\caption{Experimental results on \textbf{(a)} the impact of different precision levels in each attention block of SpikeGPT-216M, and \textbf{(b)} the performance of SpikeGPT-216M under weight quantization considering different combinations of precision levels across attention blocks~\cite{Ref_Putra_QSLM_arXiv26}.}
\label{Fig_QSLM}
\vspace{-0.2cm}
\end{figure}

Several SLMs have been proposed in the literature (e.g., SpikeBERT, SpikingBERT, SNN-BERT, SpikeLM, SpikeLLM, and SpikeGPT), and they aim at achieving high performance~\cite{Ref_Putra_QSLM_arXiv26}.
Hence, the optimization methods for SLMs have not been extensively explored. 
To this end, a quantization method based on block-wise sensitivity analysis, called QSLM, is proposed in~\cite{Ref_Putra_QSLM_arXiv26}.
Specifically, the analysis in QSLM exposes that the attention blocks are typically less sensitive than the input and output blocks when their weights are quantized; see Fig.~\ref{Fig_QSLM}(a).
Hence, a hierarchy-aware mixed-precision approach is employed to appropriately compress each block based on its sensitivity level.
Experimental results show that, this technique leads to competitive performance with significant memory savings from the baseline model (i.e., SpikeGPT-216M~\cite{Ref_Chu_SpikeGPT_TMLR24}); see Fig.~\ref{Fig_QSLM}(b).
For the SST-2 dataset, the quantized model achieves 84.4\% accuracy, comparable to 85.7\% accuracy from the baseline model, while saving 85\% of memory footprint; see~\circled{2}.
Meanwhile, for the WikiText-2 dataset, the quantized model achieves perplexity score of 24.6, comparable to 26.5 from the baseline model, while saving 68.7\% of memory footprint; see~\circled{3}.

\smallskip
\subsubsection{\textbf{Spiking Vision Transformers (SViTs)}}
Several SViTs have been proposed in the literature, such as Spikformer, Spike-Driven Transformer (SDT), and Spike-Driven Transformer v2 (SDTv2)~\cite{Ref_Putra_QSViT_IJCNN25}.
These models still focus on achieving high performance (e.g., accuracy), thus their optimization methods have not been comprehensively studied. 
Toward this, a quantization method for SViTs is proposed in~\cite{Ref_Putra_QSViT_IJCNN25}, called QSViT, leveraging layer-wise sensitivity analysis. 
Specifically, QSViT investigates the impact of block-wise quantization in the state-of-the-art SViTs on accuracy, identifies the highest and lowest precision levels across blocks that lead to acceptable accuracy, leverages this setting to explore the quantized model candidates, then selects the quantized model that achieves the highest acceptable accuracy.
QSViT effectively saves 22.75\% of memory footprint and reduces 21\% of power consumption, while maintaining high accuracy as compared to the baseline (i.e., SDTv2~\cite{Ref_Yao_SpikeDrivenTransformer2_ICLR24}).

\section{Conclusion}

The use of MFMs is growing rapidly, thereby ensuring their high energy efficiency, effective functionality, and security, is important. 
To address such design challenges, we propose a multi-layered methodology for accelerating MFMs through a design and optimization pipeline, that systematically integrates effective hardware and software techniques. 
The case studies highlight the effectiveness of our methodology in addressing design challenges in MFM optimization and deployment for high-performance and energy-efficient MFM-based systems.
In summary, our work contributes to the advancements of novel hardware and software techniques for accelerating MFMs in an integrated methodology.

\section*{Acknowledgment}

This work was partially supported by the NYUAD Center for CyberSecurity (CCS), funded by Tamkeen under the NYUAD Research Institute Award G1104, as well as partially supported by the NYUAD Center for Artificial Intelligence and Robotics (CAIR), funded by Tamkeen under the NYUAD Research Institute Award CG010.
 
\end{spacing}

\begin{spacing}{0.894}
\bibliographystyle{IEEEtran}
\bibliography{bibliography}
\end{spacing}

\end{document}